%% file: camera-ready-latex-2026.tex
\documentclass[letterpaper]{article} 
\usepackage{aaai2026}  
\usepackage{times}  
\usepackage{helvet}  
\usepackage{courier}  
\usepackage[hyphens]{url}  
\usepackage{graphicx} 
\usepackage{natbib}  
\usepackage{caption} 
\usepackage{amsmath}
\usepackage{amssymb}
\usepackage{bm} 
\usepackage{algorithm}
\usepackage{algorithmicx}
\usepackage{algpseudocode}
\usepackage{cite} 
\usepackage{marvosym}

\usepackage{mathtools}
\usepackage{booktabs}
\usepackage{subcaption}
\usepackage{cuted}

\newcommand{\pixelsota}{%
\begin{table}[t]
\centering
\small
\setlength{\tabcolsep}{4pt} 
\renewcommand{\arraystretch}{1.1}

\begin{tabular}{@{}l c c@{}} 
\toprule
\textbf{Model} & \textbf{FID}$\downarrow$ & \textbf{sFID}$\downarrow$ \\ 
\midrule
\multicolumn{3}{l}{\textit{Diffusion Models / Flow Matching}} \\ 
EDM~\cite{karras2022elucidating}      & 1.36 & --   \\
iDDPM~\cite{nichol2021improved}       & 2.92 & 3.79 \\
ADM~\cite{dhariwal2021diffusion} & 2.09 & 4.29 \\
\midrule
\multicolumn{3}{l}{\textit{Generative Adversarial Networks (GANs)}} \\ 
IC-GAN~\cite{casanova2021instance}   & 6.70 & --   \\
BigGAN~\cite{brock2018large}         & 4.06 & 3.96 \\
\midrule
\multicolumn{3}{l}{\textit{Consistency Models (CMs)}} \\ 
CD (LPIPS)~\cite{song2023consistency} & 4.70 & --   \\
iCT-deep~\cite{song2023improved}      & 3.25 & --   \\
\midrule
\multicolumn{3}{l}{\textit{Normalizing Flows}} \\ 
TARFlow~\cite{zhai2024tarflow}\textsuperscript{\textdagger} & 4.21 & 5.34 \\
\quad + \textbf{R-REPA (Ours)} & \textbf{3.69} & \textbf{4.34} \\
\bottomrule
\end{tabular}
\caption{Image generation results on class-conditional ImageNet 64$\times$64. We report FID and sFID with 50K samples. \textsuperscript{\textdagger}Result obtained using their officially released codebase.}
\label{tab:fid_imagenet64_cond}
\end{table}
}

\newcommand{\latentsotaSingleCol}{%
\begin{table}[t]
\centering
\small
\setlength{\tabcolsep}{5pt} 
\renewcommand{\arraystretch}{1.1}

\begin{tabular}{@{}l r r r@{}} 
\toprule
\textbf{Model}  & \textbf{FID}$\downarrow$ & \textbf{sFID}$\downarrow$ & \textbf{IS}$\uparrow$ \\
\midrule
\multicolumn{4}{l}{\textit{Diffusion Models}} \\ 
ADM~\citep{dhariwal2021diffusion}     & 4.59 & 5.25   & 186.70 \\
CDM~\citep{ho2022cascaded}            & 4.88 & --   & 158.71 \\
LDM-4~\citep{rombach2022high}         & 3.60 & 7.51 & 247.67 \\
DiT~\citep{peebles2023scalable}  & 2.27 & 4.60 & 278.24 \\
SiT~\citep{ma2024sit}                 & 2.06 & 4.50   & 270.30 \\
\midrule
\multicolumn{4}{l}{\textit{Autoregressive (discrete)}} \\ 
RQ-Trans.~\citep{lee2022autoregressive}  & 3.80  & -- & 323.7   \\
LlamaGen-3B~\citep{sun2024autoregressive}  & 2.18 & --   & 263.33 \\
VAR~\citep{tian2024visual}                & 1.73 & --& 350.2 \\
\midrule
\multicolumn{4}{l}{\textit{Autoregressive (continuous)}} \\ 
MAR-AR~\citep{li2024autoregressive}         & 4.69 & --   & 244.6 \\
MAR~\citep{li2024autoregressive}          & 1.55 & --   & 303.7  \\
DART~\citep{gu2024dart}                   & 3.82 & --   & 263.8  \\
\midrule
\multicolumn{4}{l}{\textit{Normalizing Flow}} \\ 
Latent-TARFlow  & 5.15 & 6.78 & 243.49 \\
\quad \textbf{+R-REPA (Ours)} & \textbf{4.95} & \textbf{6.89} & \textbf{234.99} \\
\quad \textbf{+Patch Size 1 (Ours)} & \textbf{4.18} & \textbf{4.96} & \textbf{240.8}\\
\bottomrule
\end{tabular}
\caption{Class-conditional generation on ImageNet 256$\times$256. We report FID, sFID, and IS with 50K samples. Lower is better for $\downarrow$, higher is better for $\uparrow$.}
\label{tb:imagenet_singlecol}
\end{table}
}

\newcommand{\ablation}{%
\begin{table}[t]
\centering
\small 
\setlength{\tabcolsep}{2.5pt} 
\renewcommand{\arraystretch}{1.15} 

\begin{tabular}{@{}l c c r r r r@{}}
\toprule
\textbf{Type} & \textbf{Blocks} & \textbf{Layers} & \textbf{FID}$\downarrow$ & \textbf{sFID}$\downarrow$ & \textbf{IS}$\uparrow$ & \textbf{Acc.}(\%)$\uparrow$ \\
\midrule
\multicolumn{3}{l}{TARFlow~\cite{zhai2024tarflow}}   & 12.91 & 33.79    & 36.62 & 37.43     \\
\midrule
\multicolumn{7}{l}{\textit{Setup Group 1: Alignment applied to \textbf{All} blocks}} \\
Forward       & All   & 2     & 12.25 & 37.97 & 40.85 & 46.97 \\
Detach        & All   & 2     & 12.19 & 34.31 & 41.98 & 49.06 \\
Reverse        & All   & 2     & 12.21 & 33.80 & 42.08 & 49.91 \\
\midrule
\multicolumn{7}{l}{\textit{Setup Group 2: Alignment applied to selected \textbf{2} blocks}} \\
Forward       & 1 \& 2     & 2     & 12.67 & 39.99 & 41.11 & 61.16 \\
Detach        & 1 \& 2    & 2     & 12.73 & 33.84 & 37.81 & \textbf{61.63} \\
Detach        & 7 \& 8     & 2     & 12.12 & 34.00 & 41.18 & 55.14 \\
Reverse       & 7 \& 8     & 2     & 11.93 & 33.78 & 40.90 & 55.21 \\
\midrule
\multicolumn{7}{l}{\textit{Setup Group 3: Ablation of alignment layers}} \\
Reverse       & 7 \& 8     & 2     & 11.93 & 33.78 & 40.90 & 55.21 \\
Reverse       & 7 \& 8     & 4     & 11.84 & \textbf{33.61} & \textbf{46.06} & 58.91 \\
Reverse       & 7 \& 8     & 6     & \textbf{11.71} & 33.68 & 44.31 & 57.35 \\
\bottomrule
\end{tabular}
\caption{Ablation study of our proposed alignment method on ImageNet 64$\times$64, evaluated at 400k training iterations. We vary the alignment \textit{Type} and which blocks and layers to apply it to. Best results for each metric are in \textbf{bold}.}
\label{tab:ablation}
\end{table}
}

\newcommand{\trainingdynamics}{%
\begin{table}[t]
\centering
\small
\setlength{\tabcolsep}{4pt}
\renewcommand{\arraystretch}{1.15}

\begin{tabular}{@{}lcccc@{}}
\toprule
\textbf{Model} & \textbf{Res.} & \textbf{Iter.} & \textbf{FID-4096}$\downarrow$ & \textbf{Acc.}(\%)$\uparrow$ \\
\midrule
TARFlow & 64 & 1M   & 11.76 & 39.97 \\
\quad \textbf{+R-REPA (Ours)} & 64 & \textbf{400K} & \textbf{11.71} & \textbf{57.76} \\
\quad \textbf{+R-REPA (Ours)} & 64 & \textbf{600K} & \textbf{11.53} & \textbf{57.75} \\
\quad \textbf{+R-REPA (Ours)} & 64 & \textbf{800K} & \textbf{11.48} & \textbf{57.65} \\
\textbf{\quad +R-REPA (Ours)} & 64 & \textbf{1M} & \textbf{11.25} & \textbf{57.02} \\
\midrule
Latent-TARFlow & 256 & 400K & 13.82 & 45.97 \\
Latent-TARFlow            & 256 & 1M   & 13.05 & 40.22 \\
\quad \textbf{+R-REPA (Ours)} & 256 & \textbf{400K} & \textbf{13.26} & \textbf{57.85} \\
\textbf{\quad +R-REPA (Ours)} & 256 & \textbf{1M} & \textbf{12.79} & \textbf{56.24} \\
\bottomrule
\end{tabular}
\caption{Training progress comparison. Our method, \textbf{+R-REPA}, consistently outperforms the vanilla TARFlow baselines across different checkpoints on both ImageNet 64$\times$64 and 256$\times$256 resolutions. }
\label{tab:training_dynamics}
\end{table}%
}
\usepackage{newfloat}
\usepackage{listings}
\DeclareCaptionStyle{ruled}{labelfont=normalfont,labelsep=colon,strut=off} 
\floatstyle{ruled}
\newfloat{listing}{tb}{lst}{}
\floatname{listing}{Listing}
\title{Flowing Backwards: Improving Normalizing Flows \\ via Reverse Representation Alignment}
\author{
    Yang Chen\textsuperscript{\rm 1,2},
    Xiaowei Xu\textsuperscript{\rm 2},
    Shuai Wang\textsuperscript{\rm 1},
    Chenhui Zhu\textsuperscript{\rm 1,2},
    Ruxue Wen\textsuperscript{\rm 2},\\
    Xubin Li\textsuperscript{\rm 2},
    Tiezheng Ge\textsuperscript{\rm 2},
    Limin Wang\textsuperscript{\rm 1,3,\Letter}
}
\affiliations{
    \textsuperscript{\rm 1}State Key Laboratory for Novel Software Technology, Nanjing University\\
    \textsuperscript{\rm 2}Alibaba Group, \textsuperscript{\rm 3}Shanghai AI Lab

    \Letter~Corresponding author: lmwang@nju.edu.cn \\
    Code: \url{https://github.com/MCG-NJU/FlowBack}
}

\usepackage{bibentry}

\begin{document}

\twocolumn[{%
\renewcommand\twocolumn[1][]{#1}%
\maketitle
\includegraphics[width=0.98\linewidth]{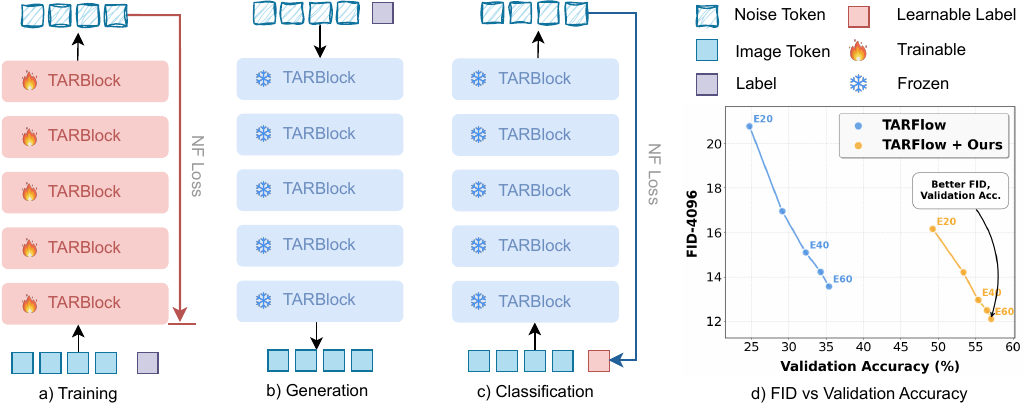}
\captionof{figure}{Take TARFlow as a representative NF. (a) Training process maps images to a noise distribution. (b) The reverse pass generates images. (c) Optimizing a label token by NF loss to classify. (d) The FID-Accuracy plot demonstrates that our representation alignment improves both generation quality and classification performance.}
\label{fig:motivation}
\vspace{1em}
}]

\newcommand\blfootnote[1]{%
  \begingroup
  \renewcommand\thefootnote{}\footnote{#1}%
  \addtocounter{footnote}{-1}%
  \endgroup
}
\begin{abstract}
Normalizing Flows (NFs) are a class of generative models distinguished by a mathematically invertible architecture, where the forward pass transforms data into a latent space for density estimation, and the reverse pass generates new samples from this space. This characteristic creates an intrinsic synergy between representation learning and data generation. However, the generative quality of standard NFs is limited by poor semantic representations from log-likelihood optimization. To remedy this, we propose a novel alignment strategy that creatively leverages the invertibility of NFs: instead of regularizing the forward pass, we align the intermediate features of the generative (reverse) pass with representations from a powerful vision foundation model, demonstrating superior effectiveness over naive alignment. We also introduce a novel training-free, test-time optimization algorithm for classification, which provides a more intrinsic evaluation of the NF's embedded semantic knowledge. Comprehensive experiments demonstrate that our approach accelerates the training of NFs by over 3.3$\times$, while simultaneously delivering significant improvements in both generative quality and classification accuracy. New state-of-the-art results for NFs are established on ImageNet 64$\times$64 and 256$\times$256.
\end{abstract}



\section{Introduction}
\label{sec:intro}

Normalizing Flows (NFs) represent a distinct class of generative models, characterized by their exact mathematical invertibility~\cite{pmlr-v37-rezende15, dinh2014nice, dinh2016density}. This property defines their rigid, dual-pathway architecture: a forward pass transforms data into a simple latent distribution for exact log-likelihood optimization, while a mathematically precise reverse pass generates data from that same latent space (Figure~\ref{fig:motivation}a,b). This structure implies the synergy of NFs between data generation and representation learning, where the two are truly two sides of the same coin.

This inherent synergy suggests a clear path toward enhancing the generative capabilities of NFs: by improving the quality of their learned representations. However, this potential remains largely underexploited. Standard NFs, optimized solely for log-likelihood on the forward pass, often fail to learn semantically meaningful features, which in turn limits their generative quality. The model's rigid adherence to the likelihood objective prevents it from fully realizing the benefits of its own architectural duality.

This line of inquiry is particularly relevant given recent findings that actively improving a model's representational quality can enhance its generative capabilities. For instance, a notable method, REPA~\cite{repa}, regularized the internal features of diffusion models against a strong, pretrained visual encoder. This `representation-first' strategy yielded significant gains in both training efficiency and generation quality, demonstrating the effectiveness of leveraging high-quality external guidance.

Inspired by REPA, our work aims to adapt and extend it to the unique context of NFs. We ask: \textbf{how can the invertible structure of NFs be leveraged to effectively synergize representation learning and generation?} By utilizing the reverse generative pathway for alignment—a strategy uniquely enabled by NFs—we move beyond direct application to explore new alignment strategies within this model class. This forms the core motivation for our work.

In this paper, we first need a way to probe the intrinsic discriminative abilities of a given NF. Departing from the standard linear-probe protocol, we introduce a novel \textbf{training-free, test-time optimization} algorithm, visually depicted in Figure~\ref{fig:motivation}c. Instead of training an auxiliary classifier, our method directly leverages the model's own loss landscape to perform classification, providing a more direct measure of its embedded semantic knowledge. Our initial evaluation using this framework confirms a critical weakness: standard NFs, despite their generative prowess, exhibit poor discriminative performance.

To address this deficiency, we propose reverse representation alignment, a novel strategy built upon two core components: representation alignment and the exploitation of the model's reverse pass. We term the act of operating on the generative pathway \textit{`flowing backwards'}. In this process, our method directly enforces semantic consistency by aligning the intermediate features during the true generative pass ($\mathbf{z}$-to-$\mathbf{x}$), a concept uniquely enabled by the invertible architecture of NFs.

As conceptually illustrated in Figure~\ref{fig:motivation}d, our comprehensive experiments show that this reverse alignment strategy is remarkably effective. It not only yields substantial improvements in generative quality as measured by FID but also, when assessed with our test-time method, unlocks a dramatic increase in classification accuracy. In summary, our main contributions are:
\begin{itemize}
    \item We design and systematically evaluate several alignment strategies for NFs, culminating in our proposed \textbf{reverse representation alignment} (R-REPA) method that uniquely leverages the model's invertibility.
    \item We propose a novel training-free, test-time optimization method for NF-based classification (Figure~\ref{fig:motivation}c), serving as both a diagnostic tool and a more intrinsic evaluation metric.
    \item We demonstrate empirically that our approach significantly enhances both the generative and discriminative performance of NFs (Figure~\ref{fig:motivation}d), establishing a new state-of-the-art for synergizing these two capabilities.
\end{itemize}

\section{Related Work}
\label{sec:related_work}
\paragraph{Normalizing Flows.}
NFs are exact likelihood-based generative models~\cite{dinh2014nice, pmlr-v37-rezende15}. While historically surpassed by diffusion models~\cite{ma2024sit,flowdcn,DDT,pixnerd} in sample quality, recent work has revitalized the field. TARFlow~\cite{zhai2024tarflow} introduces a Transformer-based architecture that achieves state-of-the-art likelihoods and generates samples with quality comparable to diffusion models. JetFormer~\cite{tschannen2024jetformer} and FARMER~\cite{farmer} leverage an NF as a core, jointly trained component within a unified autoregressive model for high-fidelity joint image-text generation, eliminating the need for pre-trained autoencoders. Our concurrent work STARFlow~\cite{starflow} successfully scales NFs in terms of both model capacity and task complexity. These works demonstrate the renewed potential of NFs when integrated with modern architectures.

\paragraph{Representation Alignment for Generation.}
\label{par:rep_align}
A recent paradigm for accelerating generative model training is Representation Alignment, which leverages features from pretrained vision foundation models (VFMs) as guidance. The seminal \textbf{REPA}~\cite{repa} introduced a loss to align the internal hidden states of a denoising network with VFM features, drastically improving convergence and final sample quality. This simple yet powerful principle has proven highly effective and was quickly extended. Subsequent work has used it to enable stable end-to-end training of the entire latent diffusion pipeline ~\cite{lee2024repae}, to improve the VAE's latent space directly ~\cite{zheng2024lightningdit}, and to adapt the alignment strategy to other backbones like U-Nets ~\cite{tian2024urepa}, demonstrating its broad utility.

\begin{figure*}
    \centering
    \includegraphics[width=0.9\linewidth]{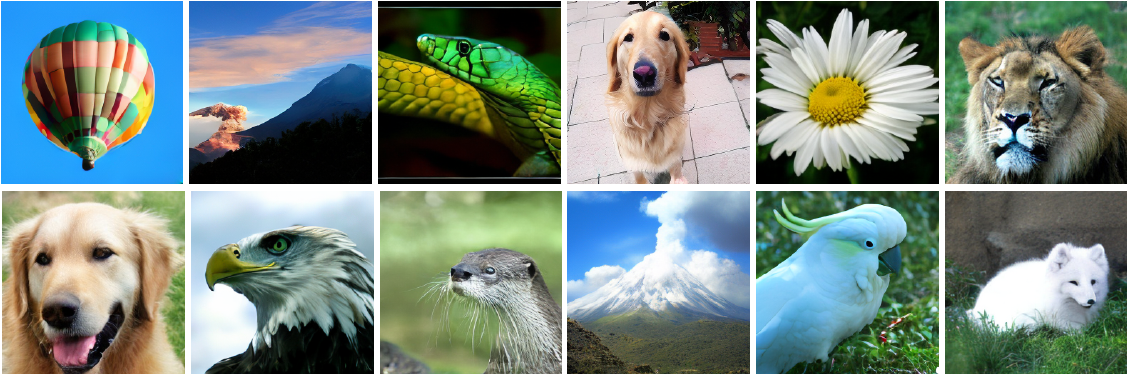}
    \caption{Selected Samples on ImageNet 256 $\times$ 256 from L-TARFlow + R-REPA. We use classifier-free guidance equal to 2.0.}
    \label{fig:vis}
\end{figure*}

\section{Preliminary: Normalizing Flow and TARFlow}

Normalizing Flows (NFs)~\cite{pmlr-v37-rezende15, dinh2014nice, dinh2016density} represent a class of generative models based on exact likelihood built on the formula for change of variables. Given a continuous data distribution $p_{\text{data}}$ for inputs $\mathbf{x} \in \mathbb{R}^D$, a Normalizing Flow learns a parametric, invertible transformation $f_\theta: \mathbb{R}^D \to \mathbb{R}^D$. This function maps complex data $\mathbf{x}$ to a simple, tractable base distribution $p_0$ (e.g., a standard Gaussian), often referred to as the noise or latent space. The model is trained by maximizing the log-likelihood (MLE) of the data:
\begin{equation}
\label{eq:nf_mle}
\begin{split}
    \max_{\theta} \mathbb{E}_{\mathbf{x} \sim p_{\text{data}}} \Big[  \log p_0(f_\theta(\mathbf{x})) 
     + \log \left| \det \frac{\partial f_\theta(\mathbf{x})}{\partial \mathbf{x}} \right| \Big],
\end{split}
\end{equation}
where the first term encourages mapping data points to high-density regions of the prior $p_0$, while the log-determinant of the Jacobian term penalizes excessive local volume shrinkage, ensuring the transformation remains bijective. A generative model is obtained by the inverse transformation $f_\theta^{-1}$, with a sampling procedure $\mathbf{x} = f_\theta^{-1}(\mathbf{z})$, where $\mathbf{z} \sim p_0(\mathbf{z})$.

A prominent and computationally efficient variant is the Autoregressive Flow (AF)~\cite{kingma2016improved, papamakarios2017masked}. In an AF, the transformation $f_\theta$ is directly defined by a pair of parameter-generating functions $(\mu_\theta, \sigma_\theta)$, which specify an element-wise affine map. Crucially, these functions are architecturally constrained to be autoregressive: the parameters for each dimension $d$ are computed using only the preceding input dimensions $\mathbf{x}_{<d}$. The forward (encoding) and inverse (sampling) processes for each dimension $d \in [1, D]$ are:
\begin{equation}
\label{eq:af_transform}
\begin{aligned}
    \mathbf{z}_d &= (\mathbf{x}_d - \mu_\theta(\mathbf{x}_{<d})) \odot \sigma_\theta(\mathbf{x}_{<d})^{-1}, \\
    \mathbf{x}_d &= \mu_\theta(\mathbf{x}_{<d}) + \sigma_\theta(\mathbf{x}_{<d}) \odot \mathbf{z}_d.
\end{aligned}
\end{equation}
This autoregressive structure ensures the Jacobian of $f_\theta$ is triangular, greatly simplifying the log-determinant term in Eq.~\ref{eq:nf_mle} to a simple sum: $-\sum_{d=1}^D \log \sigma_\theta(\mathbf{x}_{<d})$.

Recently, TARFlow~\cite{zhai2024tarflow} was introduced as a high-performance NF architecture for image data, building upon the AF framework. TARFlow is constructed by stacking multiple Transformer AutoRegressive Blocks (TARBlocks), $\mathbf{z} = f^T_\theta \circ \dots \circ f^1_\theta(\mathbf{x})$, where each block $f^t_\theta$ processes its input using a different autoregressive ordering $\pi^t$. By alternating these orderings, the stacked blocks can capture dependencies across all dimensions. The parameters $\mu$ and $\sigma$ for each block are modeled using causal Transformer layers. Assuming the final output $\mathbf{z} = \mathbf{x}^T$ follows a standard Gaussian prior, the end-to-end training objective becomes:
\begin{equation}
\label{eq:tarflow_mle}
\begin{split}
    \max_{\theta} \mathbb{E}_{\mathbf{x} \sim p_{\text{data}}} \Big[  -\frac{1}{2} \|\mathbf{z}\|_2^2 
     - \sum_{t=1}^T \sum_{d=1}^D \log \sigma^t_\theta(\mathbf{x}^{t-1}_{\pi^t_{<d}}) \Big],
\end{split}
\end{equation}
where $\mathbf{x}^t = f^t_\theta(\mathbf{x}^{t-1})$ and $\mathbf{x}^0 = \mathbf{x}$. Additionally, TARFlow employs noise augmented training and score-base denoising to improve the modeling capability.

\begin{algorithm}[tb]
\caption{Training-Free Classification via Single-Step Gradient.}
\label{alg:training_free_cls}
\begin{algorithmic}[1] 
    \Require Image $\mathbf{x}$, pre-trained model $f_\theta$, class embeddings $\mathbf{E} \in \mathbb{R}^{K \times D_{\text{emb}}}$.
    \Ensure Predicted class label $y_{\text{pred}}$.
    
    \State Initialize logits $\bm{\lambda} \leftarrow \mathbf{0} \in \mathbb{R}^K$.
    \State Compute weighted class embedding:
    \State $\mathbf{p} \leftarrow \text{softmax}(\bm{\lambda})$
    \State $\mathbf{e}_{\text{eff}} \leftarrow \mathbf{p}^T \mathbf{E}$
    
    \State Compute the log-likelihood score:
    \State $\mathcal{L}(\bm{\lambda}) \leftarrow \log p(\mathbf{x} \mid \mathbf{e}_{\text{eff}}; \theta)$
    
    \State Compute the gradient with respect to the logits:
    \State $\mathbf{g} \leftarrow \nabla_{\bm{\lambda}} \mathcal{L}(\bm{\lambda})$
    
    \State Predict the class corresponding to the largest gradient component:
    \State $y_{\text{pred}} \leftarrow \operatorname*{argmax}_{k} (\mathbf{g})_k$
    
    \State \Return $y_{\text{pred}}$
\end{algorithmic}
\end{algorithm}
\section{Methodology}
In this section, we present our proposed methods for enhancing and leveraging Normalizing Flows. We begin by introducing a novel, training-free classification algorithm that directly utilizes the learned probability density from class-conditional NFs. Subsequently, we describe a representation alignment algorithm specifically designed to improve the discriminative quality of the latent representations within NFs. Finally, to address the computational challenges of modeling high-dimensional data, we adapt the TARFlow architecture to operate within a compressed latent space, enabling efficient generation.
\subsection{Training-Free Classification with NFs}
\label{subsec:training_free_cls}
\begin{figure*}[t] 
    \centering 
    \includegraphics[width=0.9\linewidth]{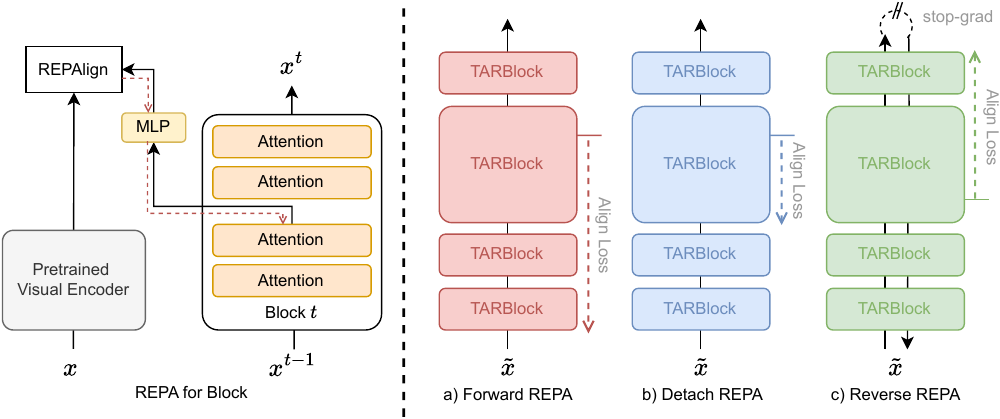} 
    \caption{An overview of our Representation Alignment (REPA) mechanism. \textbf{Left:} Intermediate features from a TARFlow block are projected by an MLP and aligned with features from a pre-trained visual encoder. \textbf{Right:} The three gradient backpropagation strategies explored: (a) Forward REPA (F-REPA), updating all preceding blocks; (b) Detach REPA (D-REPA), updating only the current block; and (c) Reverse REPA (R-REPA), which leverages the \textit{inverse (generative) computational graph} to update all subsequent blocks. While we depict alignment at a single location for clarity, this mechanism can be applied concurrently across multiple layers.} 
    \label{fig:repa} 
\end{figure*}

We introduce a novel, training-free classification algorithm that leverages the density estimation capabilities of a pre-trained class-conditional TARFlow model. Instead of training a separate classifier, our method reframes classification as an inference-time optimization problem. The core idea is to find the class label $y$ that maximizes the conditional log-likelihood $\log p(\mathbf{x}|y; \theta)$ for a given input image $\mathbf{x}$, where $\theta$ are the frozen parameters of the generative model.

We achieve this by estimating the gradient of the log-likelihood with respect to a``soft'' class conditioning. This is done by first defining a set of classification logits, $\bm{\lambda} \in \mathbb{R}^K$, which control a weighted average of the model's class embeddings $\mathbf{E}$. The gradient of the log-likelihood is then computed with respect to these logits at their initialization point. The class with the largest gradient component is selected as the prediction, as it indicates the direction of greatest increase in likelihood. The entire process requires only a single forward and backward pass through the model and is detailed in Algorithm~\ref{alg:training_free_cls}.

\subsection{Representation Alignment}
\label{subsec:feature_alignment}
While the MLE objective excels at density modeling, the learned intermediate features of a NF are not inherently optimized to be semantically meaningful. To address this, we introduce a feature alignment mechanism that injects high-level semantic guidance from a powerful, pre-trained vision model into the TARFlow's generative process.

As illustrated in Figure~\ref{fig:repa}~(left), we use a pre-trained, frozen vision encoder $\Phi(\cdot)$ to extract a semantic representation $\mathbf{v} = \Phi(\mathbf{x}) \in \mathbb{R}^{P \times D}$ from an input image $\mathbf{x}$, where $P$ is the number of patches and $D$ is the embedding dimension. The objective is to align TARFlow's intermediate features with this target representation $\mathbf{y}$.

Let $\mathbf{h}^{(t,l)}$ be the feature map from layer $l$ of block $t$ within TARFlow. We project these features into the semantic space using a learnable head, $\text{Proj}_{\phi}$ (a simple MLP). The alignment loss then maximizes the patch-wise similarity between the projected and target features:

\begin{equation}
\label{eq:align_loss_single}
\mathcal{L}_{\text{align}}^{(t,l)}(\theta, \phi) \coloneqq -\frac{1}{P} \sum_{p=1}^{P} \text{sim} \left( \mathbf{v}^{[p]}, \left[\text{Proj}_{\phi}\left(\mathbf{h}^{(t,l)}\right)\right]^{[p]} \right),
\end{equation}
where $p$ is the patch index and $\text{sim}(\cdot, \cdot)$ is a similarity function, such as cosine similarity. This alignment can be flexibly applied to any set of layers $\mathcal{A} = \{(t_1, l_1), \dots\}$.

Crucially, we explore three distinct strategies for backpropagating the gradient of this alignment loss, each manipulating the computational graph to control how the parameters are updated by the alignment loss. These strategies are visualized in Figure~\ref{fig:repa} (right).

\paragraph{Forward Strategy.} As the most direct approach, this strategy involves backpropagating the gradient of $\mathcal{L}_{\text{align}}^{(t,l)}$ through the forward computational graph. As illustrated in Figure~\ref{fig:repa} (a), this updates both the projector $\phi$ and all parameters of the TARFlow layers preceding layer $(t,l)$.

\paragraph{Detach Strategy.} This strategy draws an analogy to diffusion models, treating each TARFlow block as a network operating at a specific timestep $t$. To isolate the alignment process to this single `timestep', we detach the input to the block. Consequently, the gradient only updates the parameters within that block (i.e., $\theta_t$) and the projector $\phi$, preventing any influence on preceding blocks (Figure~\ref{fig:repa}, b).

\paragraph{Reverse Strategy.}
This novel strategy fundamentally alters the update mechanism by leveraging the computational graph of the \textit{reverse (generative) process}. Specifically, we first compute the latent variable $\mathbf{z} = f_\theta(\mathbf{x})$ via the forward pass and then detach it. A new computational graph is then constructed by executing the inverse flow $f_\theta^{-1}$, starting from the detached latent $\mathbf{z}_{\text{detached}}$. The alignment loss is computed within this inverse pass. Crucially, backpropagation from this loss occurs entirely on the generative graph, inherently confining gradient updates to the parameters of layers subsequent to the alignment layer $(t,l)$ (relative to the original forward pass). Figure~\ref{fig:repa} (c) conceptualizes this, showing how the stop-gradient on $\mathbf{z}$ reroutes the gradient path exclusively through the generative pathway.

\paragraph{Final Loss Formulation.}
The total training objective is a weighted sum of the NF loss and the averaged alignment losses from all chosen layers in the set $\mathcal{A}$:
\begin{equation}
\label{eq:total_loss}
\mathcal{L}_{\text{total}} = \mathcal{L}_{\text{NF}} + \lambda_{\text{align}} \left( \frac{1}{|\mathcal{A}|} \sum_{(t,l) \in \mathcal{A}} \mathcal{L}_{\text{align}}^{(t,l)}(\theta, \phi) \right),
\end{equation}
where $\lambda_{\text{align}}$ is a hyperparameter balancing the two terms. The gradient computation for $\mathcal{L}_{\text{align}}$ follows one of the three strategies outlined above.


\begin{figure}[t] 
    \centering 
    \includegraphics[width=\linewidth]{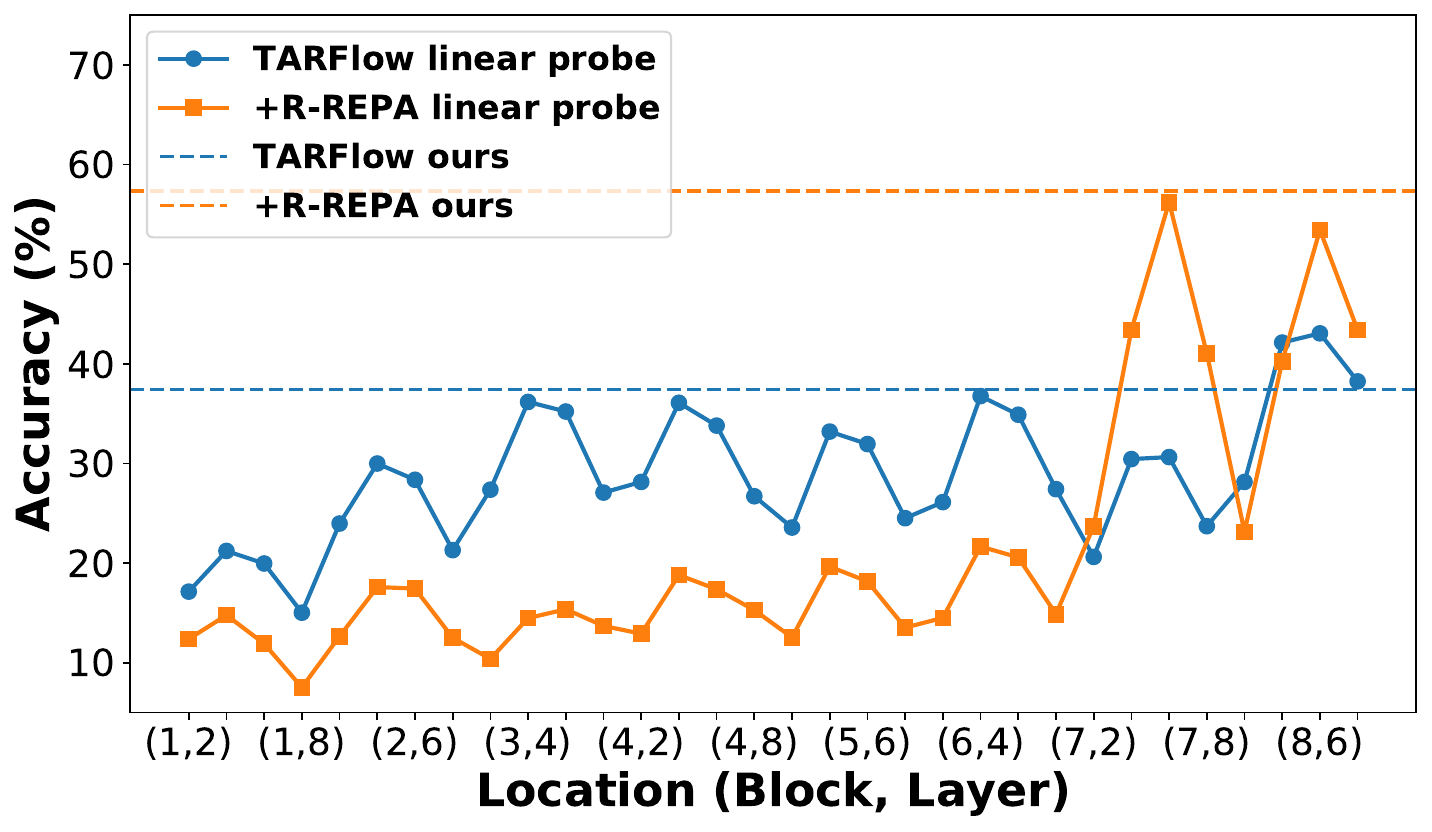} 
    \caption{Validation of our proposed classification metric against the standard linear probing protocol. The plot shows classification accuracy of our metric and standard linear probing.} 
    \label{fig:metric} 
\end{figure}

\subsection{Moving to Latent Space}

To scale our method to high-resolution synthesis, we transition from modeling pixels directly to modeling the latent space of a pre-trained Variational Autoencoder (VAE). This established strategy allows us to offload the task of low-level perceptual compression. Our primary work thus focuses on the core challenge: applying the powerful density estimation and refinement techniques of TARFlow to the compact and semantically-rich latent codes.

The training process is adapted to this latent space. For a clean latent vector $x$ obtained from the VAE encoder, our NF model, $f_{\theta}$, is trained on a noisy version $\tilde{x}$:
\begin{equation}
\tilde{x} = x + \epsilon, \quad \text{where } \epsilon \sim \mathcal{N}(0, \sigma^2 I).
\end{equation}
The model thus learns to estimate the density $p_{\theta}(\tilde{x})$.

The generative process mirrors this by first sampling from this learned noisy distribution and then applying a denoising step. Specifically, a noisy latent sample $\tilde{x}$ is generated via the inverse transformation, $\tilde{x} = f_{\theta}^{-1}(z)$, where $z \sim \mathcal{N}(0,I)$ is a sample from the base distribution. This sample is subsequently refined using the score-based denoising procedure:
\begin{equation}
\hat{x} = \tilde{x} + \sigma^2 \nabla_{\tilde{x}} \log p_{\theta}(\tilde{x}).
\end{equation}
Finally, the refined latent vector $\hat{x}$ is decoded into the final image, effectively scaling the precise likelihood modeling and powerful sample refinement of TARFlow to the high-resolution domain.

\section{Experiments}
\label{sec:experiments}
\paragraph{Dataset and Task.}
We conduct our class-conditional image generation and classification experiments on the ImageNet-1K dataset~\cite{deng2009imagenet}. Our models are trained exclusively on the training set 
and evaluated at two distinct resolutions: $64 \times 64$ and $256 \times 256$.

\begin{figure*}[tb]
    \centering 

    \begin{subfigure}{0.33\linewidth}
        \includegraphics[width=\linewidth]{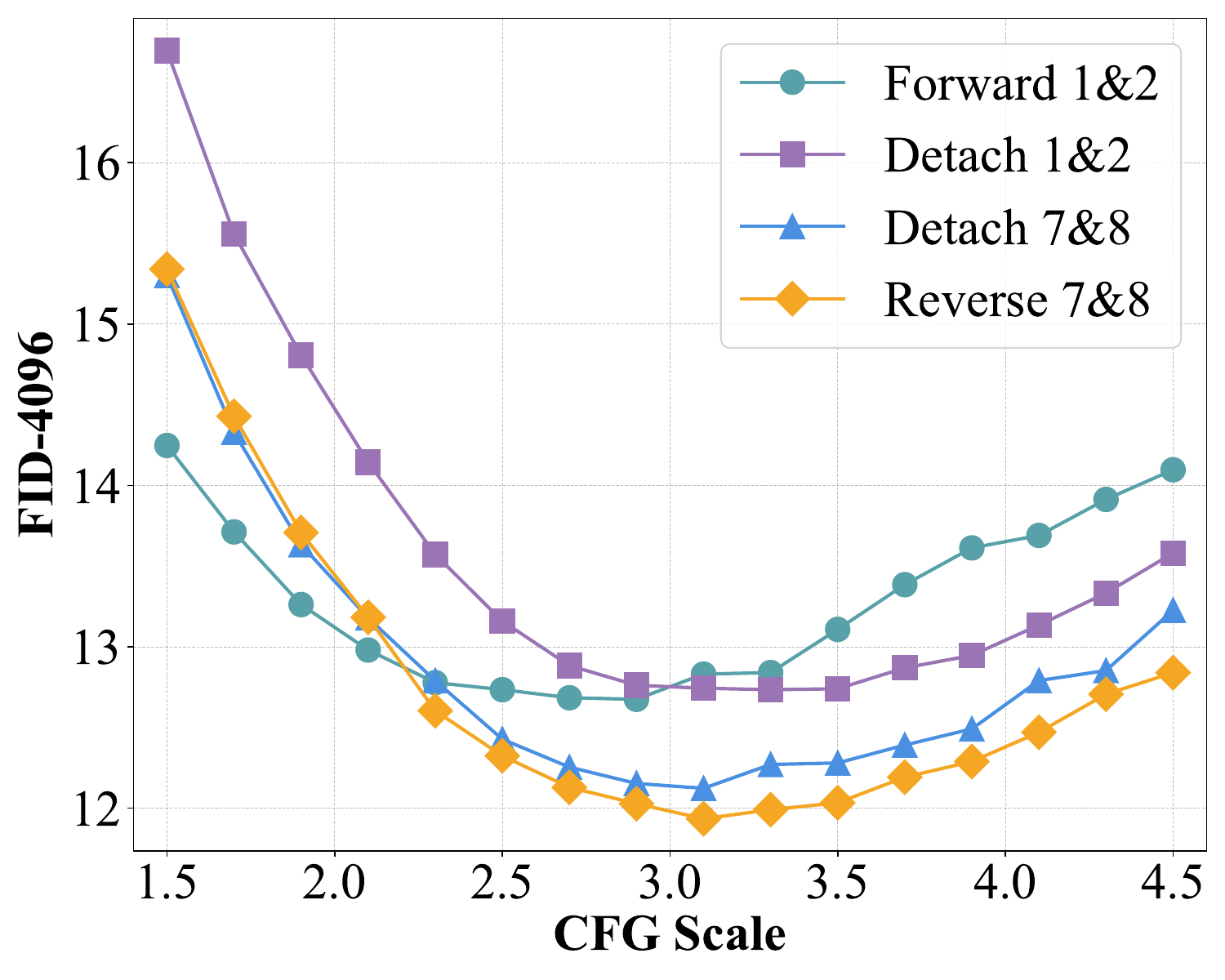}
        \label{fig:suba}
    \end{subfigure}\hfill 
    \begin{subfigure}{0.33\linewidth}
        \includegraphics[width=\linewidth]{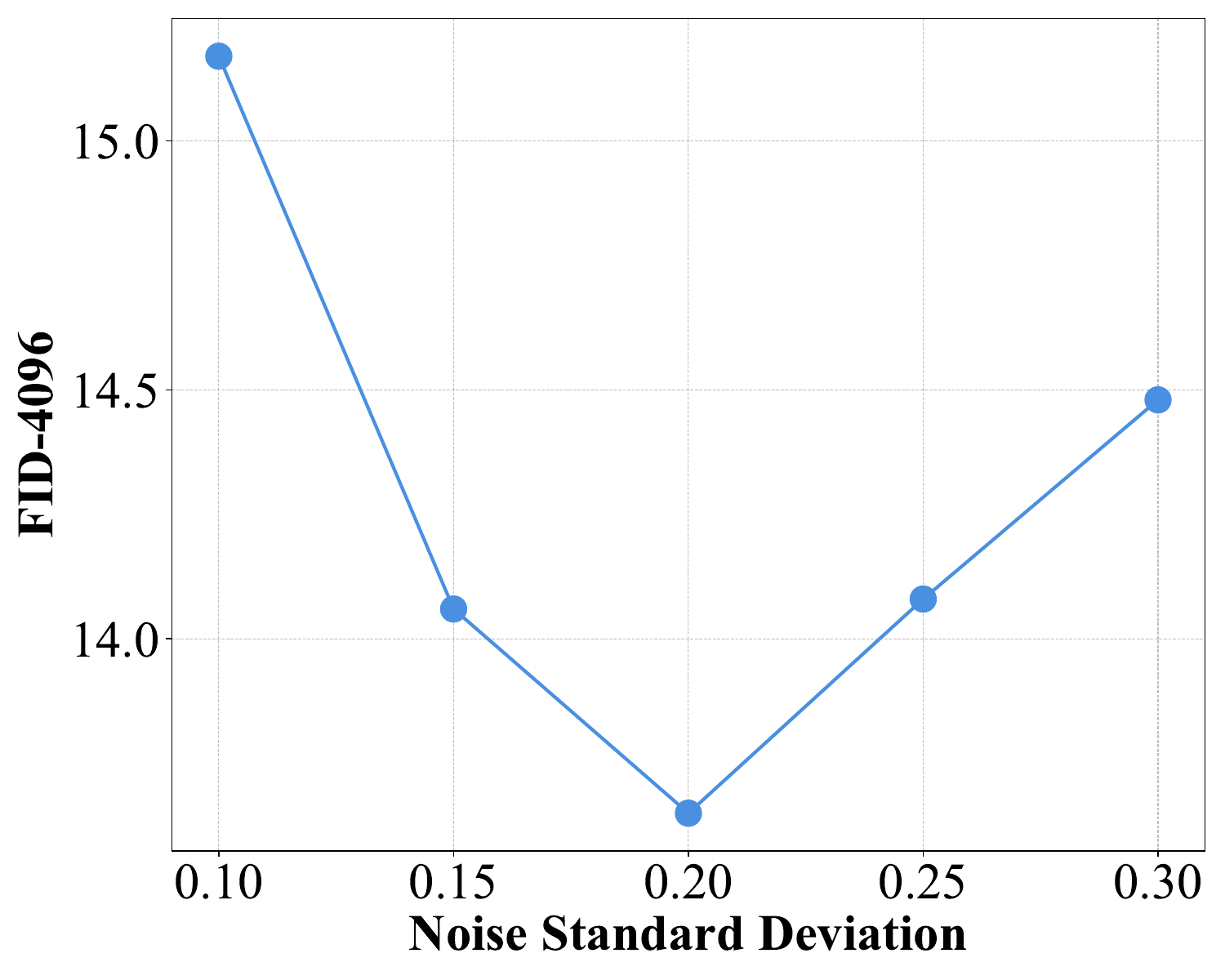}
        \label{fig:subb}
    \end{subfigure}
    \begin{subfigure}{0.33\linewidth}
        \includegraphics[width=\linewidth]{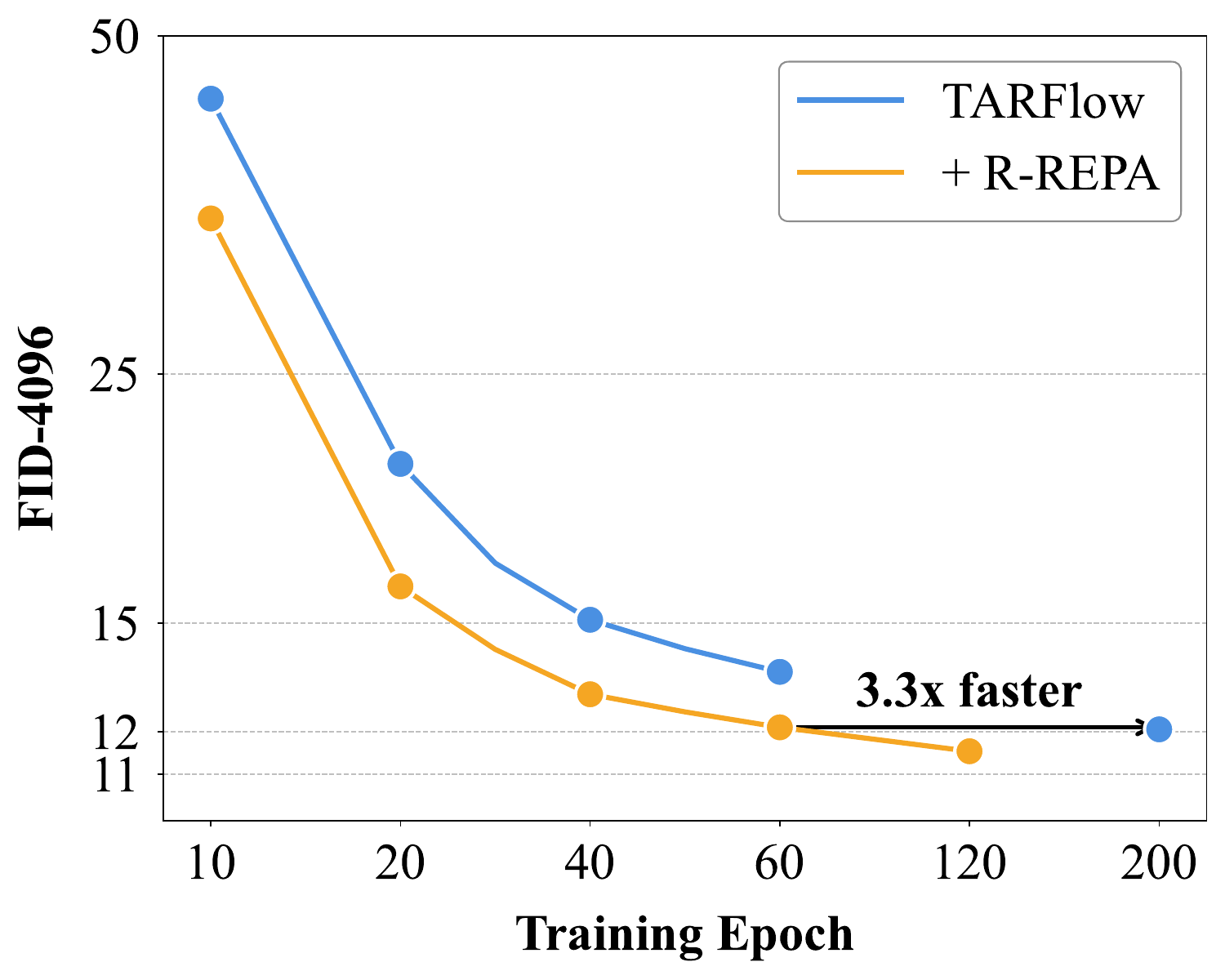}
        \label{fig:subc}
    \end{subfigure}\hfill 
    \caption{Hyperparameters Ablations and Training Convergence. 
\textbf{Left:} CFG search results on ImageNet 64$\times$64. The Reverse REPA strategy applied to later model blocks yields the best performance across various CFG scales.
\textbf{Center:} Ablation of noise standard deviation on latent space. We identify an optimal noise standard deviation of 0.20.
\textbf{Right:} Reverse REPA improves sample fidelity and accelerates training convergence by 3.3$\times$ on ImageNet 64$\times$64.}
    \label{fig:three_images_in_one_col}
\end{figure*}

\paragraph{Evaluation Metrics.}
To assess the performance of our generative model, we employ a standard suite of metrics to measure sample fidelity and diversity:  Fr\'echet inception distance (FID)~\cite{heusel2017gans}, sFID~\cite{nash2021generating}, and Inception Score (IS)~\cite{salimans2016improved}. Following TARFlow, we sample 4096 images for evaluation. Besides, we report results at the optimal CFG scale for each model, determined via a grid search detailed in Figure~\ref{fig:three_images_in_one_col}. 

To evaluate the discriminative ability, we measure the classification accuracy on the ImageNet-1K validation set. This is achieved using our proposed test-time optimization classification. Figure~\ref{fig:metric} confirms the validity of our approach. We compare our single-score evaluation (dashed lines) with the layer-specific results from standard linear probing (solid lines). While linear probing produces a wide range of outcomes depending on the layer, our method's score consistently reflects the overall performance ceiling for both TARFlow and +R-REPA models. This demonstrates that our method is a valid and efficient alternative to the expensive process of layer-by-layer linear evaluation.

\paragraph{Implementation Details.}
For experiments at the $64 \times 64$ resolution, our model architecture strictly adheres to the design of TARFlow~\cite{zhai2024tarflow}. Specifically, the model is composed of 8 TARBlocks. Each block, in turn, contains 8 layers of causal attention. The channel dimension is set to 1024, and the model operates on non-overlapping image patches of size $4 \times 4$.

For the higher resolution of $256 \times 256$, we first leverage a pre-trained VAE-ft-EMA~\cite{esser2021taming} to compress images into a lower-dimensional latent space. Our generative model then operates on this latent representation. The transformer architecture is enhanced with two key components: Rotary Position Embeddings (RoPE)~\cite{rope} and the SwiGLU activation function~\cite{llama1, llama2}. To maintain consistent patch number with 64 $\times$ 64, we use a patch size of $2 \times 2$. All other hyperparameters, including the number of TARBlocks (8), layers per block (8), and channel dimension (1024), are kept consistent with the $64 \times 64$ configuration.

\subsection{Ablation Studies}
\label{sec:ablation}
\ablation
In this section, we conduct a series of ablation studies to systematically determine our optimal model configuration. We first investigate the core design of the REPA mechanism (its backpropagation strategy, location, and depth), and then analyze key hyperparameters for training and sampling. Unless otherwise specified, all ablations are performed on ImageNet 64$\times$64 at 400k training iterations.

\paragraph{Backpropagation Strategy.}
As shown in Table~\ref{tab:ablation}, the choice of backpropagation strategy proves to be critical. We find the Forward strategy consistently degrades sFID (e.g., from 33.79 to 37.97). We hypothesize that this is because its unconstrained gradient flow creates a tension between the MLE objective and the alignment loss in the model's early blocks. Forcing these foundational blocks, which are likely specialized for low-level spatial statistics, to also conform to high-level semantics proves detrimental to overall sample coherence. The `Detach' strategy improves upon this by localizing updates. However, the `Reverse' strategy is demonstrably superior, achieving a better FID (11.93 vs. 12.12) in direct comparison. By updating only the generative path ($f_\theta^{-1}$), it effectively refines synthesis without disrupting the core density model.

\paragraph{Alignment Block and Layer.}
The choice of which blocks to align creates a trade-off between generative quality and feature semantics. Aligning Blocks 7 \& 8, which are the first to operate on the latent variable $\mathbf{z}$ during the generative process, yields the best FID (11.93). Guiding these initial synthesis steps helps establish a strong high-level image structure. Conversely, aligning Blocks 1 \& 2 produces the best semantic accuracy (\textbf{61.63\%}). These blocks are the first to process the input image $\mathbf{x}$ during encoding, so aligning them directly optimizes feature extraction but harms FID by constraining the delicate final synthesis steps.

Finally, within our best-performing setup (R-REPA, aligning 7 \& 8 blocks), we found that guiding deeper layers leads to better fidelity. As shown in Group 3, the FID progressively improves from 11.93 to \textbf{11.71} as we move alignment from the 2nd to the 6th layer. This suggests that guiding more refined features closer to a block's output provides a more potent refinement signal. While an intermediate (4th) layer achieves the best IS (\textbf{46.06}), the deepest (6th) layer is the clear choice for maximizing the final image quality.


\trainingdynamics
\paragraph{Classifier-Free Guidance (CFG) Scale.}
We optimized the CFG scale, a critical hyperparameter, for each model to ensure a fair comparison. As shown in Figure~\ref{fig:three_images_in_one_col}, our final configuration (`Reverse 7\&8') achieves the lowest FID while also demonstrating robustness across a wide scale range, with an optimum near 3.1. All results are reported at their respective optimal scales.


\paragraph{Latent Noise Std.}
For the 256$\times$256 latent-space model, the standard deviation ($\sigma$) of noise added to VAE latents (Eq. 6) is crucial. We ablated this value, with results in Figure~\ref{fig:three_images_in_one_col}. A clear performance peak exists at $\sigma=0.20$, which we use for all latent-space experiments.


\subsection{Main Results}
\label{sec:main_results}
Equipped with the optimal R-REPA configuration from our ablations, we now evaluate its performance against baselines and state-of-the-art models on ImageNet at 64$\times$64 and 256$\times$256 resolutions.

\pixelsota 
\paragraph{Performance against Baselines.}
We first compare our method, R-REPA, directly against the vanilla TARFlow baselines. As shown in Table~\ref{tab:training_dynamics}, our method provides substantial improvements in both sample quality (FID) and learned feature discriminability (Accuracy).

On ImageNet 64$\times$64, our final model achieves an FID of \textbf{11.25} and an accuracy of \textbf{57.02\%}, significantly outperforming the baseline's 11.76 FID and 39.97\% accuracy. The faster convergence of discriminative accuracy over generative quality indicates that the model learns high-level semantics early in training before progressively refining fine-grained details for synthesis. This efficiency is a crucial advantage of our approach and leads to accelerated training ($\boldsymbol{3.3\times}$), as illustrated in Figure~\ref{fig:three_images_in_one_col}. Quantitatively, our model at just \textbf{400k iterations} already surpasses the fully-trained (1M iter.) baseline in both FID (11.71 vs. 11.76) and, most notably, accuracy (57.76\% vs. 39.97\%).

A similar leap is observed on the 256$\times$256 latent-space task, where FID improves from 13.05 to \textbf{12.79} and accuracy jumps from 40.22\% to \textbf{56.24\%}. This demonstrates that representation alignment not only enhances final performance but also provides a more efficient training signal, leading to superior models in significantly less time.

\paragraph{Generation on ImageNet 64$\times$64.}
As shown in Table~\ref{tab:fid_imagenet64_cond}, our R-REPA strategy delivers state-of-the-art generative performance for Normalizing Flows on the class-conditional ImageNet 64$\times$64 benchmark. Our method significantly improves upon the strong TARFlow baseline, reducing the FID from 4.21 to \textbf{3.69} and the sFID from 5.34 to \textbf{4.34}. This performance not only surpasses established GANs like BigGAN (FID 4.06) but also brings flow-based models into closer competition with powerful diffusion models such as iDDPM (FID 2.92). Crucially, this top-tier result is achieved with just \textbf{two sampling steps}, highlighting the exceptional inference efficiency of our method compared to the multi-step iterative process required by competing paradigms.

\latentsotaSingleCol 

\paragraph{Generation on ImageNet 256$\times$256.}
We further test the scalability of our approach on the challenging ImageNet 256$\times$256 benchmark by operating in the latent space of a pre-trained VAE. As presented in Table~\ref{tb:imagenet_singlecol}, our method again demonstrates remarkable effectiveness. Our optimized configuration, which combines R-REPA with an architectural adjustment to a 1x1 patch size, achieves a highly competitive FID of \textbf{4.18} and sFID of \textbf{4.96}—a substantial improvement over the baseline. Even the direct application of R-REPA provides a clear boost, reducing FID to 4.95.

Most importantly, these competitive high-resolution results are achieved while \textbf{preserving the two-step sampling efficiency}. This demonstrates that the benefits of our approach are not confined to smaller scales but are robust and scalable. By delivering high-fidelity results with a minimal computational budget, our work establishes REPA-enhanced Normalizing Flows as a compelling and highly efficient paradigm for high-resolution image synthesis.



\section{Conclusion}
\label{sec:conclusion}

In this work, we introduce R-REPA, a novel training strategy that enhances the semantic awareness of NFs. It leverages their unique invertibility to enforce semantic consistency directly on the generative ($\mathbf{z}$-to-$\mathbf{x}$) pass, thereby unlocking the powerful synergy between representation learning and generation inherent in the architecture.
The empirical results are compelling. R-REPA establishes a new state-of-the-art for NFs on ImageNet by delivering simultaneous gains in generative fidelity (FID) and classification accuracy over the strong TARFlow baseline. This accuracy gain is rigorously quantified by our novel training-free classification algorithm—a more intrinsic probe of the model’s learned semantics. Furthermore, our method demonstrates robust high-resolution scalability while also dramatically boosting training efficiency by over $3.3\times$.
Ultimately, our work establishes a powerful new principle for advancing NFs: that fostering a virtuous cycle between semantic representation and the generative process is a direct and effective route to higher fidelity.

\section*{Acknowledgements}
This work is supported by the National Key R\&D Program of China (No. 2022ZD0160900), the Natural Science Foundation of Jiangsu Province (No. BK20250009), the Collaborative Innovation Center of Novel Software Technology and Industrialization, Alibaba Group through Alibaba Innovative Research Program.

\input{appendix}

\bibliography{aaai2026}


\end{document}

%% file: appendix.tex
\section*{Appendix}
\appendix
This appendix provides supplementary material to support the main paper. We begin in Section~\ref{sec:appendix_algo} with a comprehensive overview of the algorithmic details of our proposed R-REPA method. This includes a review of the foundational TARFlow block computation, a detailed explanation of our accelerated implementation which is critical for efficient training, and an in-depth analysis of the gradient flow that highlights the fundamental differences between our reverse-pass alignment and alternative forward-pass strategies. In Section~\ref{sec:more_exp_details}, we outline our experimental setup, covering training, sampling, classification procedures, and the computational resources used. Section~\ref{sec:appendix_additional_results} presents additional ablation studies that further validate our design choices, including the selection of alignment layers and the alignment loss weight. Finally, in Section~\ref{sec:appendix_visualizations}, we offer additional visualizations to provide further qualitative insight into the performance of our model.

\section{Algorithm Details for R-REPA}
\label{sec:appendix_algo}

\subsection{Review of the TARFlow Block Computation}
\label{sec:appendix_tarflow_review}

We begin by briefly recapping the core computational process of a single TARFlow block, $f_\theta$, which forms the basis of our model. This block transforms an input tensor $\mathbf{x} \in \mathbb{R}^{B \times D \times C}$ to an output tensor $\mathbf{z}$ of the same shape, where $B$ is the batch size, $D$ is the sequence length (i.e., number of tokens), and $C$ is the channel dimension.

As described in the main paper, the transformation is autoregressive with respect to the sequence dimension. For each token index $d \in [1, D]$, the forward (encoding) and inverse (sampling) operations are defined as:
\begin{equation}
\label{eq:af_transform_appendix}
\begin{aligned}
    \mathbf{z}_d &= (\mathbf{x}_d - \mu_\theta(\mathbf{x}_{<d})) \odot \sigma_\theta(\mathbf{x}_{<d})^{-1} \\
    \mathbf{x}_d &= \mu_\theta(\mathbf{x}_{<d}) + \sigma_\theta(\mathbf{x}_{<d}) \odot \mathbf{z}_d
\end{aligned}
\end{equation}
Here, $\mathbf{x}_d$ and $\mathbf{z}_d$ denote the tensors for the $d$-th token, and $\mathbf{x}_{<d}$ represents all preceding tokens $\{\mathbf{x}_1, \ldots, \mathbf{x}_{d-1}\}$.

A key computational property, which is central to our subsequent analysis, is that the forward pass (encoding) is parallelizable because all input tokens $\{\mathbf{x}_d\}$ are known simultaneously. In contrast, the inverse pass (sampling) is inherently sequential, as generating $\mathbf{x}_d$ requires prior computation of all $\mathbf{x}_{<d}$.

\subsection{Accelerated R-REPA Implementation}
\label{sec:appendix_accel_r_repa}
\begin{algorithm}[t]
\caption{Accelerated R-REPA Training Step}
\label{alg:r-repa-accel}
\begin{algorithmic}[1]
\State \textbf{Input:} Image tensor $\mathbf{x}$, model parameters $\theta = \{\theta_t\}_{t=1}^T$, alignment layer set $\mathcal{A}$.
\State \textbf{Hyperparameters:} Alignment loss weight $\lambda_{\text{align}}$.
\State \textbf{Pre-trained Models:} Frozen vision encoder $\Phi$, learnable projector $\phi$.

\Statex \textcolor{gray}{\# --- 1. Forward Pass and Caching ---}
\State Initialize empty list for cached features: $\texttt{cached} \leftarrow [~]$
\State $\mathbf{x}^0 \leftarrow \mathbf{x}$
\For{$t = 1$ \textbf{to} $T$}
    \State $\texttt{cached.append}(\operatorname{stop\_gradient}(\mathbf{x}^{t-1}))$
    \State $\mathbf{x}^t \leftarrow f^t_\theta(\mathbf{x}^{t-1})$ \Comment{Standard forward block}
\EndFor
\State $\mathbf{z} \leftarrow \mathbf{x}^T$
\State $\mathcal{L}_{\text{NF}} \leftarrow -\log p(\mathbf{z}) - \sum_{t=1}^T \log |\det J_{f^t_\theta}|$

\Statex \textcolor{gray}{\# --- 2. Accelerated Pseudo-Reverse Pass ---}
\State Initialize alignment loss: $\mathcal{L}_{\text{align}} \leftarrow 0$
\State Target features $\mathbf{y} \leftarrow \Phi(\mathbf{x})$
\State $\mathbf{z}^T \leftarrow \operatorname{stop\_gradient}(\mathbf{z})$
\For{$t = T$ \textbf{down to} $1$}
    \State $\hat{\mathbf{x}}^{t-1} \leftarrow \texttt{cached}[t-1]$
    \State \Comment{Compute pseudo-inverse using $\mathbf{z}_{\text{rev}}^t$ and $\hat{\mathbf{x}}^{t-1}$}
    \State $\mathbf{z}^{t-1}, \{h^{(t,l)}\}_{l} \leftarrow (f^t_\theta)^{-1}_{\text{accel}}(\mathbf{z}^t, \text{cond}=\hat{\mathbf{x}}^{t-1})$
    
    \For{each layer $l \in \mathcal{A}_t$}
    \State $\mathcal{L}_{\text{align}} \leftarrow \mathcal{L}_{\text{align}} - \operatorname{sim}(\operatorname{Proj}_\phi(h_{\text{rev}}^{(t,l)}), \mathbf{y})$
    \EndFor
\EndFor

\Statex \textcolor{gray}{\# --- 3. Total Loss and Parameter Update ---}
\State $\mathcal{L}_{\text{total}} \leftarrow \mathcal{L}_{\text{NF}} + \lambda_{\text{align}} \mathcal{L}_{\text{align}}$
\State Update $\theta$ and $\phi$ using the gradient $\nabla_{\theta, \phi} \mathcal{L}_{\text{total}}$
\end{algorithmic}
\end{algorithm}

The core concept of our Reverse Representation Alignment (R-REPA) is to apply an alignment loss during the generative (reverse) pass of the TARFlow model. However, a naive implementation of the reverse pass, $\mathbf{x} = f_\theta^{-1}(\mathbf{z})$, is computationally prohibitive for training. Each inverse block $(f^t_\theta)^{-1}$ is autoregressive, meaning the computation for each token $(\mathbf{x}^{t-1})_d$ depends on the previously generated tokens $(\mathbf{x}^{t-1})_{<d}$. This inherent sequentiality makes parallel computation on GPUs inefficient and creates a severe training bottleneck.

To overcome this, we introduce an accelerated implementation specifically for calculating the R-REPA loss during training. This method cleverly utilizes features cached during the standard forward pass to break the sequential dependency of the reverse pass.

The procedure is as follows. During the standard forward pass, which computes the sequence of transformations $\mathbf{x}^0 \rightarrow \mathbf{x}^1 \rightarrow \dots \rightarrow \mathbf{x}^T = \mathbf{z}$ (where $\mathbf{x}^0 = \mathbf{x}$), we cache the input to each block, $\mathbf{x}^{t-1}$. Crucially, we detach these cached tensors from the computational graph using a `stop\_gradient' operation, denoted as $\hat{\mathbf{x}}^{t-1} = \text{stop\_gradient}(\mathbf{x}^{t-1})$.

Next, we construct a ``pseudo-reverse'' pass to compute the alignment loss. We start with the final latent variable, $\mathbf{z}^T = \text{stop\_gradient}(\mathbf{z})$. Then, for each block $t$ from $T$ down to $1$, we perform a modified inverse operation. Instead of conditioning on sequentially generated outputs, we use the corresponding pre-computed and cached tensor $\hat{\mathbf{x}}^{t-1}$ to provide the autoregressive context. This allows the inverse transformation for all tokens to be computed in parallel:
\begin{equation}
\label{eq:accel_inverse}
    \mathbf{z}^{t-1}_d = \mu^t_\theta(\hat{\mathbf{x}}^{t-1}_{<d}) + \sigma^t_\theta(\hat{\mathbf{x}}^{t-1}_{<d}) \odot \mathbf{z}^t_d
\end{equation}
While the output of this operation, $\mathbf{z}^{t-1}$, is numerically identical to the cached $\hat{\mathbf{x}}^{t-1}$ due to the block's invertibility, this formulation creates a valid computational graph. The R-REPA loss, $\mathcal{L}_{\text{align}}^{(t,l)}$, is calculated using intermediate features $\mathbf{h}_{\text{rev}}^{(t,l)}$ from this pseudo-reverse pass.

The gradient from $\mathcal{L}_{\text{align}}$ backpropagates through the parameters of block $t$ ($\theta_t$) and, via the dependency on $\mathbf{z}^t$, continues to flow to all subsequent blocks ($\theta_{t+1}, \dots, \theta_T$). The `stop\_gradient' on the conditioning tensor $\hat{\mathbf{x}}^{t-1}$ prevents gradients from affecting any preceding blocks ($\theta_1, \dots, \theta_{t-1}$). This achieves the exact update pattern required for R-REPA while transforming the reverse pass into a fully parallelizable computation. The entire process is detailed in Algorithm~\ref{alg:r-repa-accel}.

\subsection{Computational Efficiency}

Our accelerated implementation for R-REPA is highly effective in practice. Compared to a naive approach that faithfully executes the sequential autoregressive computation, our parallelized method yields a substantial speedup of approximately \textbf{50$\times$}. Furthermore, it reduces the peak memory footprint by nearly \textbf{50\%}, making the reverse-pass alignment computationally feasible for deep models.

We also provide a direct comparison of the training throughput for our final R-REPA strategy against the Forward and Detach variants. Table~\ref{tab:speed_comparison} details the performance under an identical experimental setup where two blocks are aligned (corresponding to Group 2 in Table 1 of the main paper, note that align block location does not influence speed of Detach-REPA).

\begin{table}[t]
\centering
\begin{tabular}{@{}lcc@{}}
\toprule
\textbf{Strategy} & \textbf{Aligned Blocks} & \textbf{Speed (iter/s)} \\ \midrule
Forward-REPA & 2 & 2.11 \\
Detach-REPA  & 2 & 1.93 \\
Reverse-REPA (Ours) & 2 & 1.87 \\ \bottomrule
\end{tabular}
\caption{Training throughput comparison of different REPA strategies. The accelerated R-REPA introduces only a minor overhead compared to other methods, making it highly practical. The results are tested on the same devices.}
\label{tab:speed_comparison}
\end{table}

As shown in Table~\ref{tab:speed_comparison}, the Forward strategy is the fastest, as it only requires adding a loss to the existing forward computational graph. Both the Detach and our accelerated Reverse strategies introduce a minor computational overhead due to graph manipulation (detaching inputs or constructing the pseudo-reverse pass), resulting in a slight reduction in throughput. However, this modest decrease in speed is a small price for the significant improvements in generative quality, as demonstrated in the main paper. This makes our accelerated approach a highly practical and effective training method.

\subsection{Detailed Gradient Flow Analysis for Forward-REPA and R-REPA}
\label{sec:appendix_grad_flow}

To elucidate the fundamental difference between the Forward-REPA and our proposed Reverse-REPA (R-REPA) strategies, we analyze the gradient backpropagation pathways in a simplified two-block TARFlow model. This analysis reveals how each strategy directs the alignment supervision to different parts of the model.

\paragraph{Toy Model Setup.}
Consider a model composed of two sequential TARFlow blocks, $f^1_\theta$ and $f^2_\theta$, with parameters $\theta_1$ and $\theta_2$, respectively. The forward (encoding) process is $\mathbf{x} \xrightarrow{f^1_\theta} \mathbf{x}^1 \xrightarrow{f^2_\theta} \mathbf{z}$. We apply alignment at both blocks by extracting intermediate features $\mathbf{h}^1$ from $f^1_\theta$ and $\mathbf{h}^2$ from $f^2_\theta$. The corresponding alignment losses, $\mathcal{L}_{\text{align}}^1$ and $\mathcal{L}_{\text{align}}^2$, are computed by comparing the projected features against a target $\mathbf{y} = \Phi(\mathbf{x})$ via a projector $\text{Proj}_\phi$. The total alignment loss is $\mathcal{L}_{\text{align}} = \mathcal{L}_{\text{align}}^1 + \mathcal{L}_{\text{align}}^2$.

\paragraph{Analysis of Forward-REPA.}
In the Forward strategy, the alignment losses are computed on the standard forward computational graph. The gradient flow is as follows:
\begin{itemize}
    \item \textbf{Gradient from $\mathcal{L}_{\text{align}}^2$}: The gradient $\nabla \mathcal{L}_{\text{align}}^2$ updates the projector parameters $\phi$. Since $\mathbf{h}^2$ is a function of the parameters $\theta_2$ and the input $\mathbf{x}^1$, the gradient backpropagates to update $\theta_2$. Critically, because $\mathbf{x}^1 = f^1_\theta(\mathbf{x})$, the gradient continues to flow through $\mathbf{x}^1$ to update the parameters of the preceding block, $\theta_1$. Thus, $\nabla \mathcal{L}_{\text{align}}^2$ affects $\{\phi, \theta_2, \theta_1\}$.
    \item \textbf{Gradient from $\mathcal{L}_{\text{align}}^1$}: Similarly, $\nabla \mathcal{L}_{\text{align}}^1$ updates $\phi$ and the parameters of the first block, $\theta_1$. The gradient path terminates here as it does not depend on any subsequent blocks. Thus, $\nabla \mathcal{L}_{\text{align}}^1$ affects $\{\phi, \theta_1\}$.
\end{itemize}
The core principle of Forward-REPA is that an alignment loss at a given layer updates the parameters of that layer and \textbf{all preceding layers} in the computational graph. This constitutes a form of causal correction, adjusting upstream computations to improve downstream feature representations.

\paragraph{Analysis of R-REPA.}
R-REPA fundamentally alters the gradient pathway by constructing a new computational graph based on the generative (inverse) process, as detailed in Algorithm~\ref{alg:r-repa-accel}.
\begin{itemize}
    \item \textbf{Gradient from $\mathcal{L}_{\text{align}}^2$}: This loss is computed during the pseudo-reverse pass using features from $(f^2_\theta)^{-1}_{\text{accel}}(\mathbf{z}_{\text{detached}}, \text{cond}=\hat{\mathbf{x}}^1)$. The gradient updates $\phi$ and $\theta_2$. However, the `stop\_gradient' operation on the conditioning input $\hat{\mathbf{x}}^1$ explicitly severs the computational graph, preventing the gradient from flowing back to $\theta_1$. Thus, $\nabla \mathcal{L}_{\text{align}}^2$ affects only $\{\phi, \theta_2\}$.
    \item \textbf{Gradient from $\mathcal{L}_{\text{align}}^1$}: This loss is computed using features from $(f^1_\theta)^{-1}_{\text{accel}}(\mathbf{x}_{\text{rev}}^1, \text{cond}=\hat{\mathbf{x}})$. The gradient updates $\phi$ and $\theta_1$. Crucially, the input to this operation, $\mathbf{x}_{\text{rev}}^1$, is the \textit{output} of the previous reverse step, i.e., $\mathbf{x}_{\text{rev}}^1 = (f^2_\theta)^{-1}(\dots)$. Therefore, a dependency on $\theta_2$ exists, and the gradient flows ``upward'' through the generative chain to also update $\theta_2$. Thus, $\nabla \mathcal{L}_{\text{align}}^1$ affects $\{\phi, \theta_1, \theta_2\}$.
\end{itemize}
The core principle of R-REPA is that an alignment loss at a given layer updates the parameters of that layer and \textbf{all subsequent layers} (relative to the original forward pass). This provides generative guidance, optimizing how higher-level abstract representations (from later blocks) are transformed into lower-level, semantically-aligned features (by earlier blocks).

\paragraph{Summary of Differences.}
R-REPA's reverse flow mechanism ensures that semantic guidance is injected in a manner that aligns with the model's generative direction, optimizing from abstract latents toward concrete data, which we posit is key to its enhanced performance.

\begin{table}[t]
\centering
\begin{tabular}{@{}ll@{}}
\toprule
\textbf{Hyperparameter} & \textbf{Value} \\ \midrule
Batch size & 256 \\
Optimizer & AdamW \\
Betas ($\beta_1, \beta_2$) & (0.9, 0.95) \\
Learning Rate & 1e-4 \\
Weight Decay & 1e-4 \\
EMA Decay Rate & 0.9999 \\
$\lambda_{\text{align}}$ & 0.1 \\ 
$\text{sim}(\cdot, \cdot)$ & cos. sim \\ \bottomrule
\end{tabular}
\caption{Key hyperparameters for model training.}
\label{tab:training_hyperparams}
\end{table}

\section{More Experimental Details}
\label{sec:more_exp_details}

\subsection{Training Setup}
We train our models using the AdamW optimizer and maintain an exponential moving average (EMA) of the model weights for evaluation. Consistent with TARFlow~\cite{zhai2024tarflow}, our data augmentation pipeline includes a resize, center crop and a random horizontal flip. For the R-REPA component, we follow the precedent set by prior work on representation alignment~\cite{repa}. Specifically, we employ a frozen \textbf{DINOv2-B}~\cite{oquab2023dinov2} as the pre-trained visual encoder to provide the target representations. The learnable projector, which maps TARFlow's internal features to the DINOv2 feature space, is a 3-layer MLP with SiLU activation functions and a hidden dimension of 1024. The key hyperparameters used for training are summarized in Table~\ref{tab:training_hyperparams}. These settings are consistent for imagenet 64$\times$64 and 256$\times$256. As for the VAE, we take the off-shelf VAE-ft-EMA with a downsample factor of 8 and channel 4 from huggingface~\cite{StabilityAI2022sdvaeftema}.

\begin{figure*}
    \centering
    \includegraphics[width=0.85\linewidth]{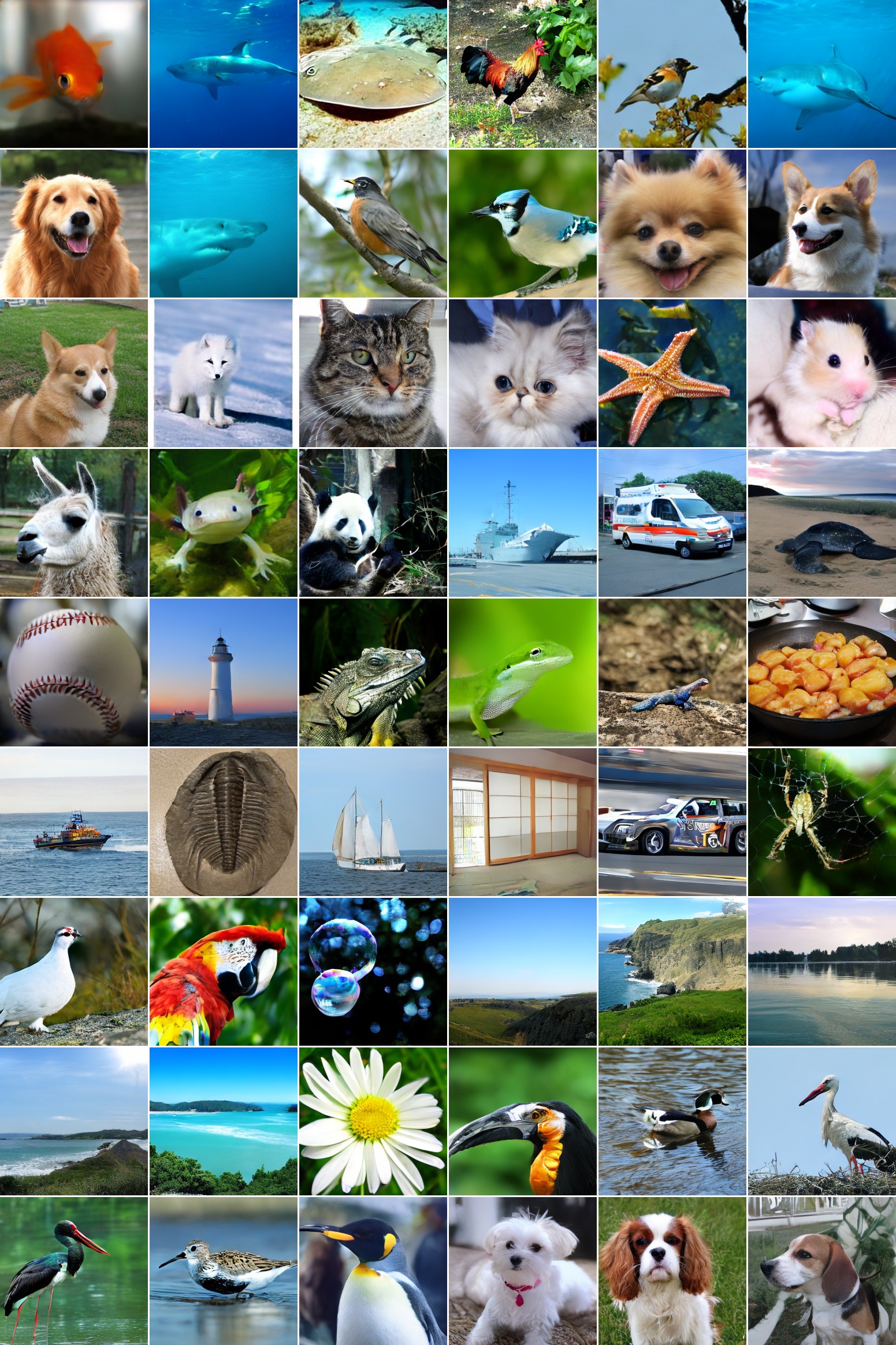}
    \caption{Additional samples on ImageNet 256 $\times$ 256 from L-TARFlow + R-REPA with PS=2, CFG=2.0.} 
    \label{fig:256}
\end{figure*}

\subsection{Sampling Setup}
For generating samples, our procedure strictly follows the methodology established in TARFlow~\cite{zhai2024tarflow}. To ensure deterministic and reproducible sampling for evaluation, which minimizes variance from stochasticity, we employ a fixed random seed. The initial noise tensor $\mathbf{z} \sim \mathcal{N}(0, \mathbf{I})$ is generated on the CPU before being transferred to the GPU for the inverse transformation process.

\subsection{Classification Setup}
Our training-free classification is performed as described in Algorithm 1 of the main paper. We investigated the sensitivity of the classification outcome to several implementation choices, including multi-step optimization instead of a single gradient step, various early stopping criteria, and different learning rates for the logits. We found the final classification prediction to be remarkably robust to these variations. This robustness stems from the core principle of the algorithm: to identify the class that corresponds to the direction of the steepest ascent in the conditional log-likelihood. A single gradient computation is sufficient to determine this direction effectively, making further optimization steps or complex heuristics largely redundant for the final prediction.

\subsection{Computational Resources}
All experiments were conducted with 8 NVIDIA GPUs. The ablation studies, which were run for 400,000 iterations, required approximately 2.5 days to complete for the ImageNet 64$\times$64 models and 3.5 days for the 256$\times$256 latent-space models. The full training for our final models, which extends for 2.5 times as many iterations (1 million total), took proportionally longer, requiring approximately 6-7 days for the 64$\times$64 models and 8-9 days for the 256$\times$256 models.

\begin{table}[t!]
\centering
\small
\begin{tabular}{lccccc}
\toprule
\textbf{Strategy} & \textbf{Blocks} & \textbf{Layer} & \textbf{FID} $\downarrow$ & \textbf{sFID} $\downarrow$ & \textbf{Acc (\%)} $\uparrow$ \\
\midrule
Detach & All & 2nd & 12.19 & 34.31 & 49.00 \\
Detach & All & 4th & 12.14 & 33.97 & 54.14 \\
Detach & All & 6th & 12.37 & 34.32 & 54.98 \\
\bottomrule
\end{tabular}
\caption{Additional ablation study of the alignment layer for the \textbf{Detach} strategy, applied to all blocks. This analysis supplements Table 1 from the main paper, showing the limited impact of layer choice in this specific setting, and further demonstrating the advantages of \textbf{Reverse} strategy. All metrics are evaluated on ImageNet 64$\times$64 at 400K iterations.}
\label{tab:ablation_detach_layer}
\end{table}

\section{Additional Experimental Results}
\label{sec:appendix_additional_results}
In this section, we provide additional ablation studies that supplement the results presented in the main paper. These experiments further justify our hyperparameter choices and reinforce the core findings of our work.

\subsection{Further Ablation on Alignment Layer for the Detach Strategy}

To further investigate the design choices presented in Table 1 of the main paper, we conducted an additional ablation study on the alignment layer depth specifically for the \textbf{Detach} strategy. In this setup, we applied representation alignment to all TARBlocks and varied the internal layer used for alignment.

As shown in Table~\ref{tab:ablation_detach_layer}, varying the alignment layer from the 2nd to the 6th results in only minor fluctuations in generative quality, with FID scores remaining within a narrow range (12.14 to 12.37). While deeper layers (4th and 6th) show a slight improvement in classification accuracy, the overall impact on sample fidelity (FID, sFID) is not substantial. This suggests that within this specific \textbf{Detach} configuration, the model's performance is not highly sensitive to the precise layer chosen for alignment. This observation validates our initial design choice in Table 1 of using the 2nd layer as a consistent setting for comparing different strategies, as it simplifies the experimental setup without significantly biasing the results for the \textbf{Detach} method.

Furthermore, it is important to highlight that even the best-performing \textbf{Detach} configuration from this ablation (FID 12.14, Acc 54.98\%) is outperformed by our proposed \textbf{Reverse} strategy (FID 11.71, Acc 57.35\% with layer 6th, as shown in Table 1 of the main paper). This comparison reinforces the superiority of leveraging the generative pathway for representation alignment.

\subsection{Ablation on Encoder Architectures}
\label{sec:encoder_ablation}

To evaluate the robustness of our proposed R-REPA method, we conduct an ablation study using various pre-trained vision encoders. Following our final setting, R-REPA is applied to layer 6 of blocks 7 and 8. We replace the default encoder(DINOv2-B) with several popular models: CLIP-B, MAE-B, and DINOv2-L. The quantitative results are presented in Table~\ref{tab:encoder_ablation}.

\begin{table}[t]
  \centering
  \begin{tabular}{lcccc}
    \toprule
    \textbf{Encoder} & \textbf{FID} $\downarrow$ & \textbf{sFID} $\downarrow$ & \textbf{IS} $\uparrow$ & \textbf{Acc} $\uparrow$ \\
    \midrule
    CLIP-B        & 12.41 & 33.99 & 41.99 & 38.99 \\
    MAE-B         & 12.32 & 34.00 & 39.69 & 37.57 \\
    DINOv2-B      & 11.71 & 33.68 & 44.31 & 57.35 \\
    DINOv2-L      & 11.74 & 33.87 & 45.00 & 58.80 \\
    \bottomrule
  \end{tabular}
  \caption{Quantitative comparison using different encoders with R-REPA. The baseline FID (12.91) represents the model without R-REPA. Arrows indicate whether a higher ($\uparrow$) or lower ($\downarrow$) value is better.}
  \label{tab:encoder_ablation}
\end{table}

The experiment highlights two critical points:

\begin{enumerate}
    \item \textbf{Robustness across encoders:} R-REPA demonstrates strong performance and robustness across all tested encoder architectures. Each configuration significantly outperforms the baseline FID of 12.91, indicating that our method is a general-purpose enhancement not limited to a specific encoder.
    
    \item \textbf{Correlation between generation and discrimination:} The results reveal a strong positive correlation between generation quality (FID) and the discriminative power of the features (Acc). Encoders that yield better accuracy, such as DINOv2, also achieve the best FID scores. This observation empirically supports our motivation of improving generation quality by enhancing the discriminative features of the underlying model.
\end{enumerate}

\subsection{Ablation on Alignment Loss Weight $\lambda_{\text{align}}$}
\begin{table}[t]
\centering
\begin{tabular}{lcc}
\toprule
\textbf{Model} & $\mathbf{\lambda_{\text{align}}}$ & \textbf{FID} $\downarrow$ \\
\midrule
Baseline & 0 & 13.82 \\
Latent-TARFlow + R-REPA & 0.1 & 13.26 \\
Latent-TARFlow + R-REPA & 0.5 & 13.74 \\
\bottomrule
\end{tabular}
\caption{Ablation study on the alignment loss weight $\lambda_{\text{align}}$ on ImageNet 256$\times$256 with 400K iterations. The results justify our choice of $\lambda_{\text{align}}=0.1$ in the main experiments.}
\label{tab:ablation_lambda}
\end{table}
We analyzed the sensitivity of R-REPA to the alignment loss weight, $\lambda_{\text{align}}$, on the ImageNet 256$\times$256 benchmark. This hyperparameter, defined in Equation 5 of the main paper, balances the standard Normalizing Flow objective ($L_{\text{NF}}$) with our proposed representation alignment loss ($L_{\text{align}}$).

Table~\ref{tab:ablation_lambda} presents the FID scores for $\lambda_{\text{align}}$ values of 0.1 and 0.5. The results clearly indicate that a smaller weight of 0.1 yields a better FID (13.26) compared to a larger weight of 0.5 (13.74). This suggests that while the alignment provides a crucial semantic signal, an overly strong alignment term can interfere with the primary density estimation task, slightly degrading the final sample fidelity. Based on this finding, we selected $\lambda_{\text{align}} = 0.1$ as the optimal value for all main experiments, a choice that is now empirically validated by this ablation.

\section{Additional Visualizations}
\label{sec:appendix_visualizations}
In this section, we provide further visual evidence of our model's generative capabilities at both high and low resolutions.

Figure~\ref{fig:256} displays additional, randomly selected samples from our \textbf{Latent-TARFlow + R-REPA} model on the ImageNet 256$\times$256 benchmark. To generate these images, we used a classifier-free guidance (CFG) scale of 2.0. The model itself operates on a \textbf{2$\times$2 patch representation} within the VAE latent space, demonstrating its capability to synthesize diverse and high-fidelity images.

\begin{figure}[t]
    \centering
    \includegraphics[width=0.8\linewidth]{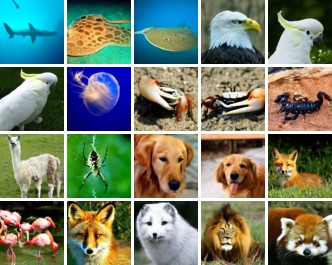}
    \caption{Samples on ImageNet 64 $\times$ 64 from TARFlow + R-REPA with PS=4, CFG=2.5.}
    \label{fig:64}
\end{figure}

In a similar vein, Figure~\ref{fig:64} presents samples for the ImageNet 64$\times$64 resolution. These were generated by our \textbf{TARFlow + R-REPA} model, which processes the input as a sequence of \textbf{4$\times$4 image patches}, employing a CFG scale of 2.5. The results confirm the model's ability to produce outputs with consistent quality and semantic coherence.

\section*{Ethical Statement}
\label{sec:ethics}

We acknowledge the ethical considerations associated with the development and application of powerful generative models. Our work, which focuses on enhancing Normalizing Flows for high-fidelity image generation, carries implications that merit careful consideration.

\paragraph{Potential for Misuse.}
Our research contributes to the advancement of high-fidelity image generation. Like all powerful generative technologies, the methods presented herein have a dual-use nature. While intended for positive applications such as creative content generation, data augmentation, and advancing fundamental machine learning research, we recognize the potential for misuse. This includes the creation of synthetic media for malicious purposes, such as generating misinformation or deceptive content. We advocate for the responsible development and deployment of generative models and encourage the broader research community to continue developing robust detection and mitigation strategies for such misuse.

\paragraph{Bias and Fairness.}
Our models are trained on the ImageNet-1K dataset and leverage features from a DINOv2 model, which was pre-trained on a large corpus of web data. It is well-documented that such large-scale datasets may contain societal biases, including demographic and stereotypical representations. Consequently, our model may inherit and potentially amplify these biases. The generated outputs may not be representative of all populations and could reflect the biases present in the training data. We urge users of our models and methods to be critically aware of these limitations and to conduct fairness assessments before deployment in any real-world application.

\paragraph{Data and Model Licensing.}
The datasets (ImageNet-1K) and pre-trained models (DINOv2, VAE-ft-EMA) used in this work are publicly available and were used in accordance with their respective licenses, which permit academic research. Our work does not involve the use of personally identifiable or sensitive private data.

\paragraph{Computational Resources and Environmental Impact.}
The training of our models required significant computational resources, as detailed in Appendix B.4. We are mindful of the environmental impact associated with large-scale machine learning experiments. A key contribution of our work is the demonstration of a more efficient training paradigm for Normalizing Flows, which accelerates convergence by over 3.3$\times$. This improvement directly contributes to reducing the overall computational cost and associated energy consumption required to achieve state-of-the-art performance, representing a step towards more sustainable AI research.

Our primary goal is to advance the scientific understanding of generative models. We release our findings to the research community to foster further innovation and to encourage a transparent and critical discussion about the capabilities and societal implications of these technologies.